\documentclass{article}

\usepackage[preprint, nonatbib]{nips_2018}

\usepackage[utf8]{inputenc} % allow utf-8 input
\usepackage[T1]{fontenc}    % use 8-bit T1 fonts
\usepackage{hyperref}       % hyperlinks
\usepackage{url}            % simple URL typesetting
\usepackage{booktabs}       % professional-quality tables
\usepackage{amsfonts}       % blackboard math symbols
\usepackage{nicefrac}       % compact symbols for 1/2, etc.
\usepackage{microtype}      % microtypography
\usepackage{graphicx}
\usepackage{amsmath}
\usepackage{float}
\usepackage{stfloats}
\usepackage{subfig}
\usepackage{amssymb}
\graphicspath{ {images/} }
\usepackage[square, numbers, sort, comma]{natbib}
\usepackage[toc,page]{appendix}

\title{A Comprehensive Comparison between Neural Style Transfer and Universal Style Transfer}

\author{
  Somshubra Majumdar\\
  Department of Computer Science\\
  University of Illinois at Chicago\\
  Chicago, IL 60607 \\
  \texttt{smajum6@uic.edu} \\
  \And
    Amlaan Bhoi \\
    Department of Computer Science \\
    University of Illinois, Chicago\\
    Chicago, IL 60607 \\
    \texttt{abhoi3@uic.edu} \\
    \And
    Ganesh Jagadeesan \\
    Department of Computer Science \\
    University of Illinois, Chicago \\
    Chicago, IL 60607 \\
    \texttt{cjagad2@uic.edu} \\
}

\begin{document}

\maketitle

\begin{abstract}
  Style transfer aims to transfer arbitrary visual styles to content images. We explore algorithms adapted from two papers that try to solve the problem of style transfer while generalizing on unseen styles or compromised visual quality. Majority of the improvements made focus on optimizing the algorithm for real-time style transfer while adapting to new styles with considerably less resources and constraints. We compare these strategies and compare how they measure up to produce visually appealing images. We explore two approaches to style transfer: \textit{neural style transfer with improvements} and \textit{universal style transfer}. We also make a comparison between the different images produced and how they can be qualitatively measured.
\end{abstract}

\section{Introduction}
Given two images, style transfer aims to transfer the style feature representation of one onto the content of the other. \textit{Convolutional neural networks} have shown to effectively learn lower level representations as well as more abstract features of an image. This means we can use CNNs for style transfer as we can preserve the style feature representations of one image and then apply it to a content image. In this paper, we first define the problem of style transfer, describe the different approaches we explore as well as their advantages and disadvantages, attempt to find evaluation measures for our results, and finally show some qualitative results.

\section{Style Transfer}
\label{style-transfer}

As discussed above, style transfer is obtained by minimizing a loss function which incorporates the semantic information of the style with the salient features of the content image. We used the VGG-16 model \cite{SimonyanZ14a} for both neural style transfer and universal style transfer, and either directly optimize the below loss function or to train a feed forward network to approximate the optimization procedure over the two losses, the \textit{content loss} and \textit{the style loss} \cite{nst}:
\begin{equation}
    L = \alpha \ || I_o - I_c ||^2_2 + \beta \ || \phi(I_o) - \phi(I_s)||^2_2
\end{equation}
Here, $I_o$, $I_c$ and $I_s$ are the feature maps from the forward pass of the VGG-16 network at certain layers that we define heuristically. The above $\alpha$ and $\beta$ are scaling weights for the two loss components, where $\alpha$ determines the strength of the \textit{Frobenius norm} between the content and generated images, and $\beta$ determines the strength of the \textit{Frobenius norm} from the \textit{feature map correlations} derived by the \textit{gram matrices} ($\phi$) between the style and generated image feature maps.

Here, the Gram Matrix $\phi$ can be computed as:
\begin{equation}
    \phi(x) = \frac{1}{HWC} \sum_{h=1}^{H} \sum_{w=1}^W x_{h,w,c} \ x_{h,w,c'}
\end{equation}
where $x$ is the feature maps of the a provided layer in the VGG-16 network, and $\phi(x)$ is proportional to the uncentered covariance of each of the channels $C_k$ in that layer, treating each image location as an independent sample.

While this objective function is sufficient, the resultant generated images are particularly noisy due to significant differences between the feature correlations of certain layers. To reduce said grain, we incorporate a regularizer, called \textit{Total Variation regularization} \cite{aly2005totalvariation}, which reduces the above problem. It can be defined as:
% \begin{equation}
%     J(x) = \int_W {L(||\triangledown_x f(x)||) dx}
% \end{equation}
$$ J(x) = \int_W L(\left \| \triangledown_x f(x))  \right \|)dx $$

More details and explanations about total variation regularization can be found in the paper by \textit{Aly et al} \cite{aly2005totalvariation}.

Finally, the objective can be defined as the minimization of the linearly scaled sum of the 3 losses described above. The values of the 2 scaling factors $(\alpha, \beta)$ are obtained from the paper by \textit{Johnson et al} \cite{johnson2016perceptual}. For the value of $\lambda$, we experimented with several values via grid search, and chose a value of \textit{8.5e{-2}} which balances the requirement of a crisp image with the noise from a grainy image.
\begin{equation}
    L = \alpha \ || I_o - I_c ||^2_2 + \beta \ || \phi(I_o) - \phi(I_s)||^2_2 + \lambda \ J(I_o)
\end{equation}
\section{Approaches}
\label{approaches}

\subsection{Neural Style Transfer}

Originally, style transfer could be achieved simply by optimizing an image initialized with Gaussian noise and minimizing the above loss function using an optimizer such as L-BFGS or Adam \cite{kingma2014adam} for a few thousand iterations. It was subjectively observed that L-BFGS obtained a much more appealing final output than Adam, although Adam was more memory efficient.

The original style transfer algorithm can be improved by using a variety of techniques discussed by \textit{Novak et al} \cite{NovakN16improving}. We incorporate a few of the proposed improvements such as utilizing all of the convolution layers of the VGG-16 to compute the overall style loss, use a geometric weighing of the style loss from each of these layers ($w_l^s = 2 ^ {(D - d(l))}$), incorporate \textit{activation shift} in the \textit{Gram Matrices} 
\begin{equation}
    \phi(x) = \frac{1}{HWC} \sum_{h=1}^{H} \sum_{w=1}^W (x-1)_{h,w,c} \ (x-1)_{h,w,c'} 
\end{equation}
and apply \textit{Chained Correlation} to determine feature correlations between adjacent layers of the network at the same spatial dimensions ($\{ \phi(x_l, x_{l-1}) \ | \ l = 2 \dots 13 \}$) where
\begin{equation}
    \phi(x, y) = \frac{1}{HWC} \sum_{h=1}^{H} \sum_{w=1}^W (x-1)_{h,w,c} \ (y-1)_{h,w,c'} 
\end{equation}
When applied to the original style transfer technique, the combination of all of these improvements significantly improves the subjective quality of the generated images.

A significant drawback of style transfer is that the feature correlations obtained from the \textit{Gram matrices} does not incorporate the color information from the original content image. This causes the generated image to have the color palette of the style image, which might not be realistic or appealing. Work done by \textit{Gatys et al.} \cite{colorpreserve} incorporates \textit{Color transform}, a method of preserving the color statistics from the content image to the generated image. While there exist two techniques, \textit{Luminance matching} and \textit{Histogram matching}, we focus primarily on \textit{Histogram matching}.

We choose this transformation so that the mean and covariance of the RGB values in the new style image $S'$ match those of $C'$. Consider $\mu_C$ and $\mu_S$ be the mean colors of the content and style image respectively, $\Sigma_C$ and $\Sigma_S$ be the pixel covariances. We then need to choose \textbf{A} and \textbf{b} such that the transform $x' = Ax + b$ yields $\mu_{S'} = \mu_C$ and $\Sigma_{S'} = \Sigma_C$, where \textit{A} is a 3x3 matrix and \textit{b} is a 3 dimensional vector. Those can be satisfied by the constraints :
$$b = \mu_C - A\mu_S$$
$$A \Sigma_S A^T = \Sigma_C$$
While there exist a family of solutions for the above problem, we can quickly find a solution to the above using 3D Color Matching formulations.  First, let the eigenvalue decomposition of a covariance matrix be $\Sigma = U \Delta U^T$. Then the matrix square root can be defined as : $\Sigma^{1/2} = U \Delta^{1/2} U^T$. Finally, the \textit{Histogram color transform} can be computed as : 
\begin{equation}
    A_{IA} = \Sigma_C^{1/2} \Sigma_S^{{-1}/2}
\end{equation}
An important extension of style transfer is the ability to mask certain regions where the transfer process should not occur. This problem is discussed in the work done by \textit{Chan et al.} \cite{ChanMaskedStyleTransfer}, which proposes the utilization of binary masks to provide the algorithm with guidance on which aspects of the content image must not be transformed. The binary mask provided is rescaled for each of the layers where the style loss is computed and the result of hadamard product of the mask with all feature maps of that layer is then used to compute the style loss. This technique allows several important extensions to style transfer, such as scaled style transfer (where the magnitude of the mask with values in the range $[0, 1]$ will determine the strength of style loss at a given position),  binary masked style transfer (where binary masks determine which of the 2 styles will be applied at a certain position) and even n-ary masked style transfer (where more than 2 styles are disambiguated using pre-determined mask values).

\subsection{Universal Style Transfer}

Universal style transfer performs style transfer by approaching the problem as an image reconstruction process coupled with feature transformation, i.e., whitening and coloring \cite{ust}. The authors in the original paper constructed an VGG-19 auto-encoder network for image reconstruction. This network was then fixed and a decoder network trained to invert the VGG-19 features to the original image.

The main difference between Universal Style Transfer and previous approaches is the introduction of the feature transformations: \textit{whitening} and \textit{coloring}. Given a pair of content image $I_c$ and style image $I_s$, the algorithm first extracts the vectorized VGG-19 feature maps $f_c \in \mathbb{R}^{C \times H_c W_c}$ and $f_s \in \mathbb{R}^{C \times H_s W_s}$ at a certain layer (e.g., Relu\_5\_1), where $H_c$, $W_c$ ($H_s$, $W_s$) are height and width of the content (style) feature, and C is the number of channels. The decoder then reconstructs the image $I_c$ given $f_c$. \\

\subsubsection{Feature Transformations}

\textbf{Whitening Transform.} The model first centers $f_c$ by subtracting its mean vector $m_c$. Then $f_c$ is transformed linearly to remove correlation between $\hat{f_c}\hat{f_c}^T=I$. This is given by: $$\hat{f_c}=E_cD_c^{-1/2}E_c^Tf_c,$$ where $D_c$ is a diagonal matrix with the eigenvalues of the covariance matrix $f_cf_c^T \in \mathbb{R}^{C \times C}$, and $E_c$ is the corresponding orthogonal matrix of eigenvectors, satisfying $f_cf_c^T \in E_cD_cE_c^T$.

\textbf{Coloring Transform.} The same centering operation is done as with Whitening Transform, but is done to the style image. We first center $f_s$ by subtracting its mean vector $m_s$, and then carry out coloring transform which is the inverse of whitening to transform $f_c$ as before to obtain $\hat{f_{cs}}$ which has the desired correlations between its feature maps ($\hat{f_{cs}}\hat{f_{cs}}^T=f_sf_s^T$),

$$\hat{f_{cs}}=E_s D_s^{1/2}E_s^T\hat{f_c},$$

where $D_s$ is a diagonal matrix with eigenvalues of covariance matrix $f_sf_s^T \in \mathbb{R}^{C \times C}$, and $E_s$ is the corresponding orthogonal matrix of eigenvectors. Finally, we re-center the $\hat{f_{cs}}$ with mean vector $m_s$ of style. When compared to histogram matching, WCT helps transfer the global color of the style image as well as salient visual patterns. After WCT, we blend $\hat{f_{cs}}$ with content feature map $f_c$ before feeding into decoder as: $$\hat{f_{cs}} = \alpha \hat{f_{cs}} + (1-\alpha)f_c,$$ where $\alpha$ serves as the style weight for controlling the transfer effect.

\subsubsection{Multi-level coarse-to-fine stylization}

Different layers of VGG networks (Relu\_X\_1) capture different levels of image structure. Higher layers capture more complicated local structures while lower layers capture more low-level information. This is due to the increasing size of receptive field and feature complexity in network hierarchy. Thus, it is more advantageous to use features from all layers instead of just the last layer. \\

WCT is applied on Relu\_5\_1 features to obtain a coarse stylized result and it is considered as the new content image to adjust features in the lower level. Experiments clearly show that higher layers capture salient patterns of style and lower levels improve the details. If we go the other way (fine-to-coarse layers), lower level information cannot be preserved.

% // TODO : Add details of the feed forward pass, the model build, the technique applied for training the model, etc

% \subsection{Other Variants of Style Transfer}

\section{Evaluation}
\label{evaluation}

Evaluating artistic style transfer is difficult. Note that the loss measures defined above for each model is based on the gram matrices, thus making the measurement and reductions in loss very much subjective to that particular image. There is no good quantitative measure to determine the overall effectiveness of a style transfer model. Thus, our primary evaluation will be \textbf{qualitative} and will rely on user's perception of effective style transfer, generally meaning how well the style has been adapted onto the content without overwhelming it.

One other aspect we can compare with other models would be \textit{speed} and \textit{efficiency}. How fast can one algorithm produce visually appealing images when compared to other algorithms. This can be a trade-off issue with \textbf{speed vs quality}. This determination needs to be made by the user.

Further, a third aspect for evaluation would be user control: how flexible a method is in adapting a user's particular requirements on the \textbf{stylization} and \textbf{sizes of images} that can be fed to the models and the \textbf{sizes of the outputs}.

\section{Results}
\label{results}

\begin{figure}[t!]
\begin{tabular}{cc}
  \includegraphics[width=65mm]{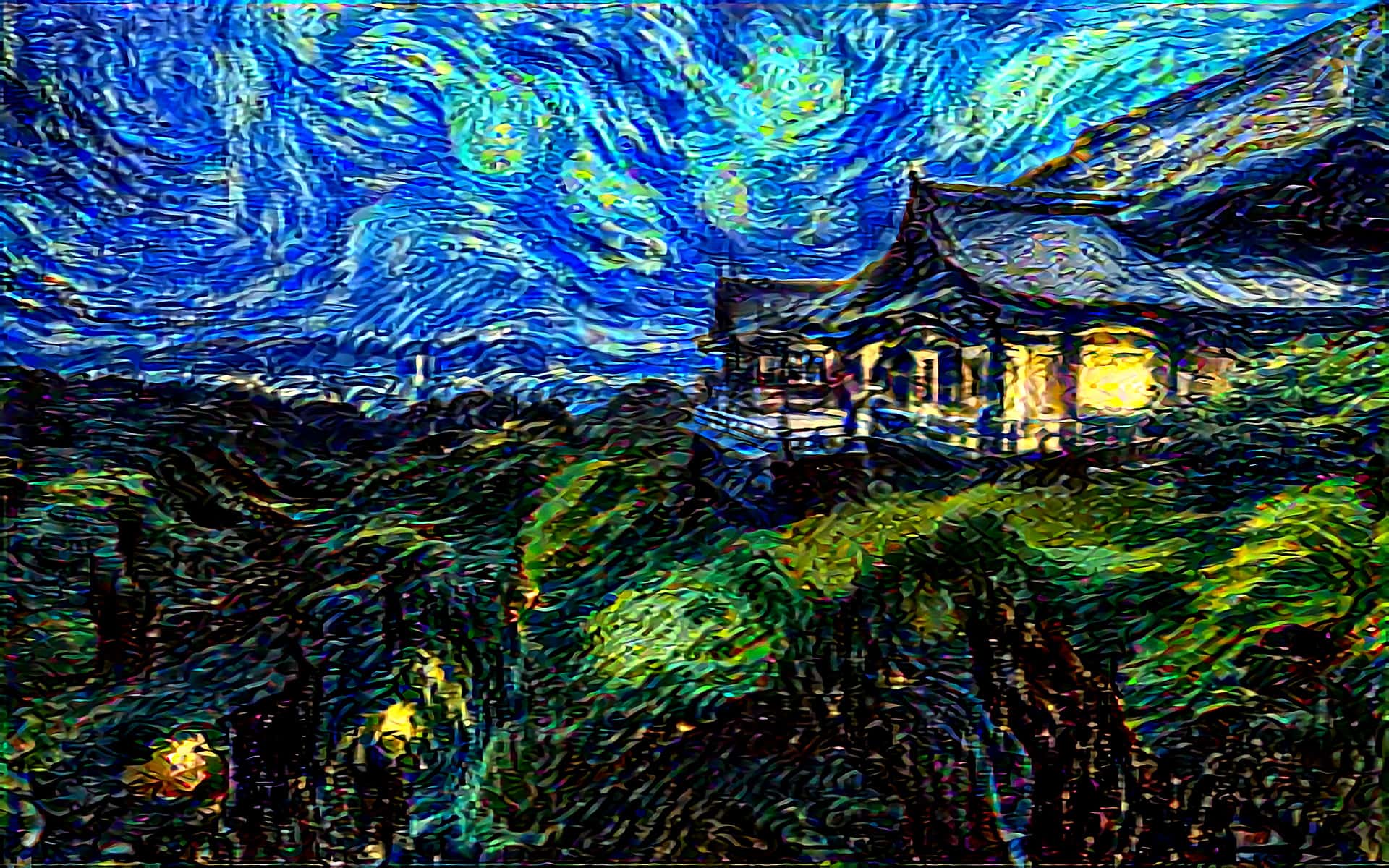} &   \includegraphics[width=65mm, height=40.5mm]{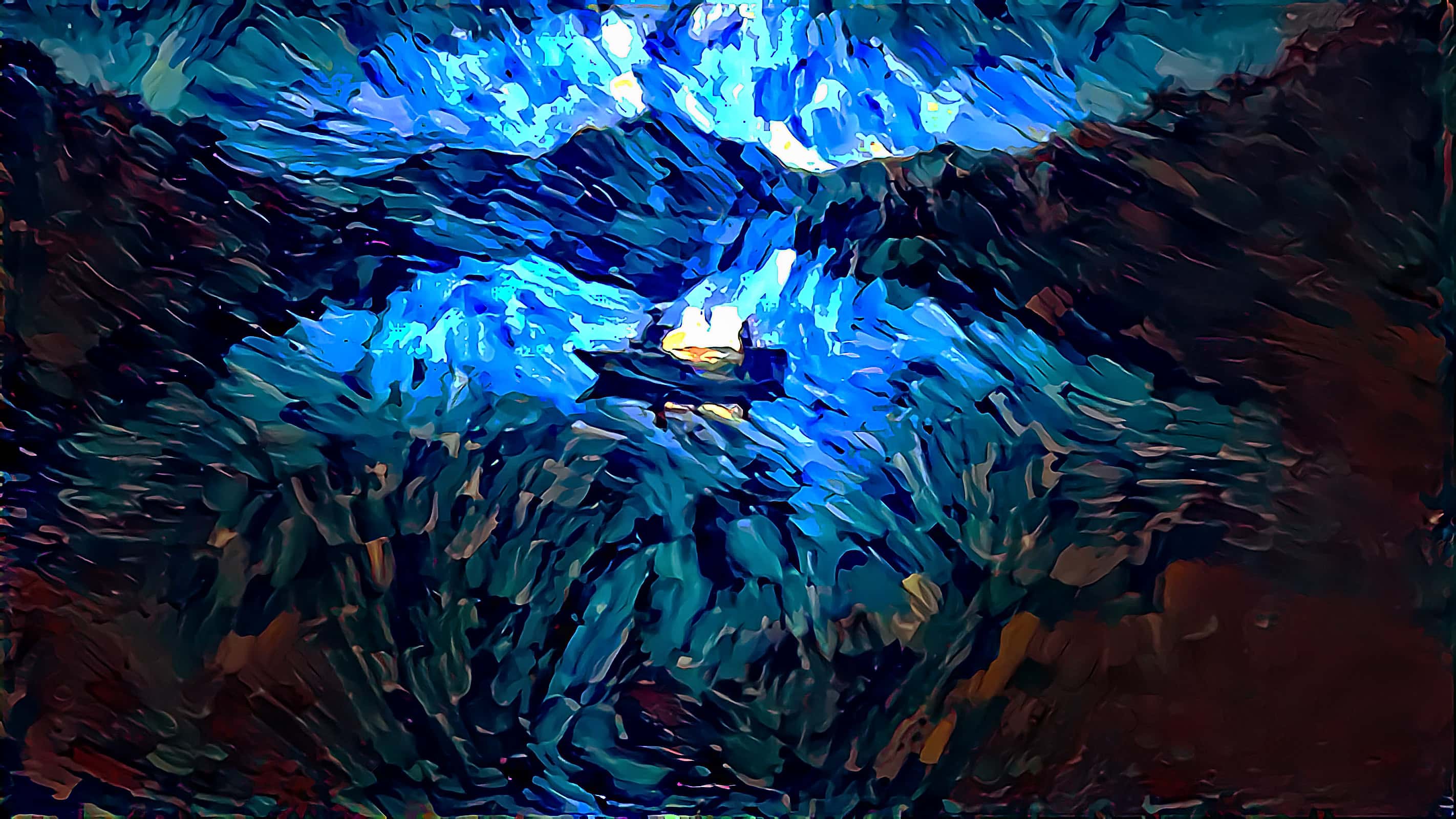} \\
(a) Japanese shrine \& Starry Night (Van Gogh) & (b) Milky Way \& Blue Strokes + Color \\[6pt]
 \includegraphics[width=65mm]{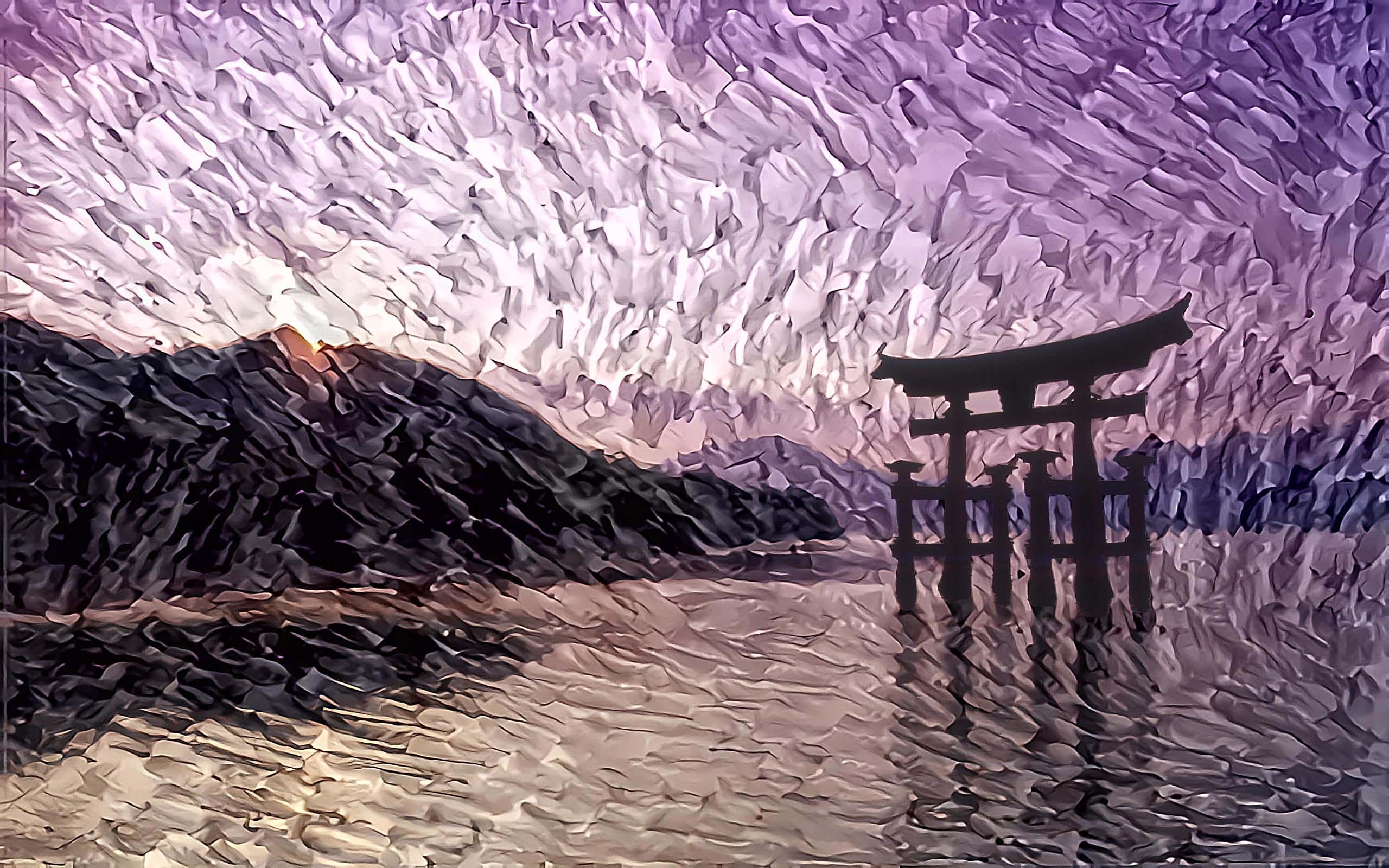} &   \includegraphics[width=65mm]{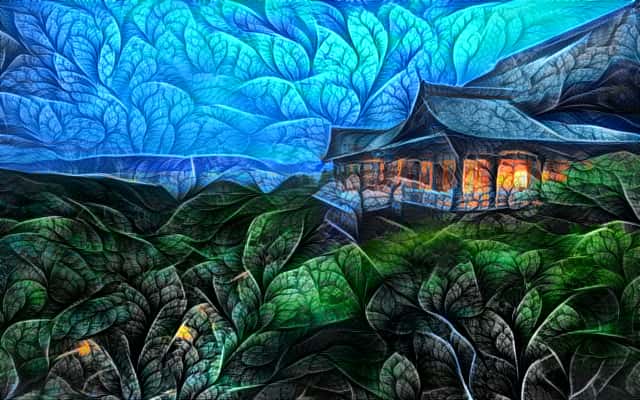} \\
(c) Itsukushima Shrine \& Blue Strokes + Color & (d) Japanese shrine \& Patterned Leaf + Color \\[6pt]
 \includegraphics[width=65mm]{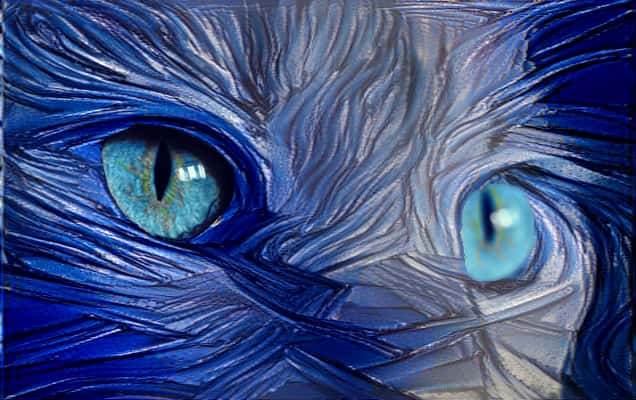} &   \includegraphics[width=65mm]{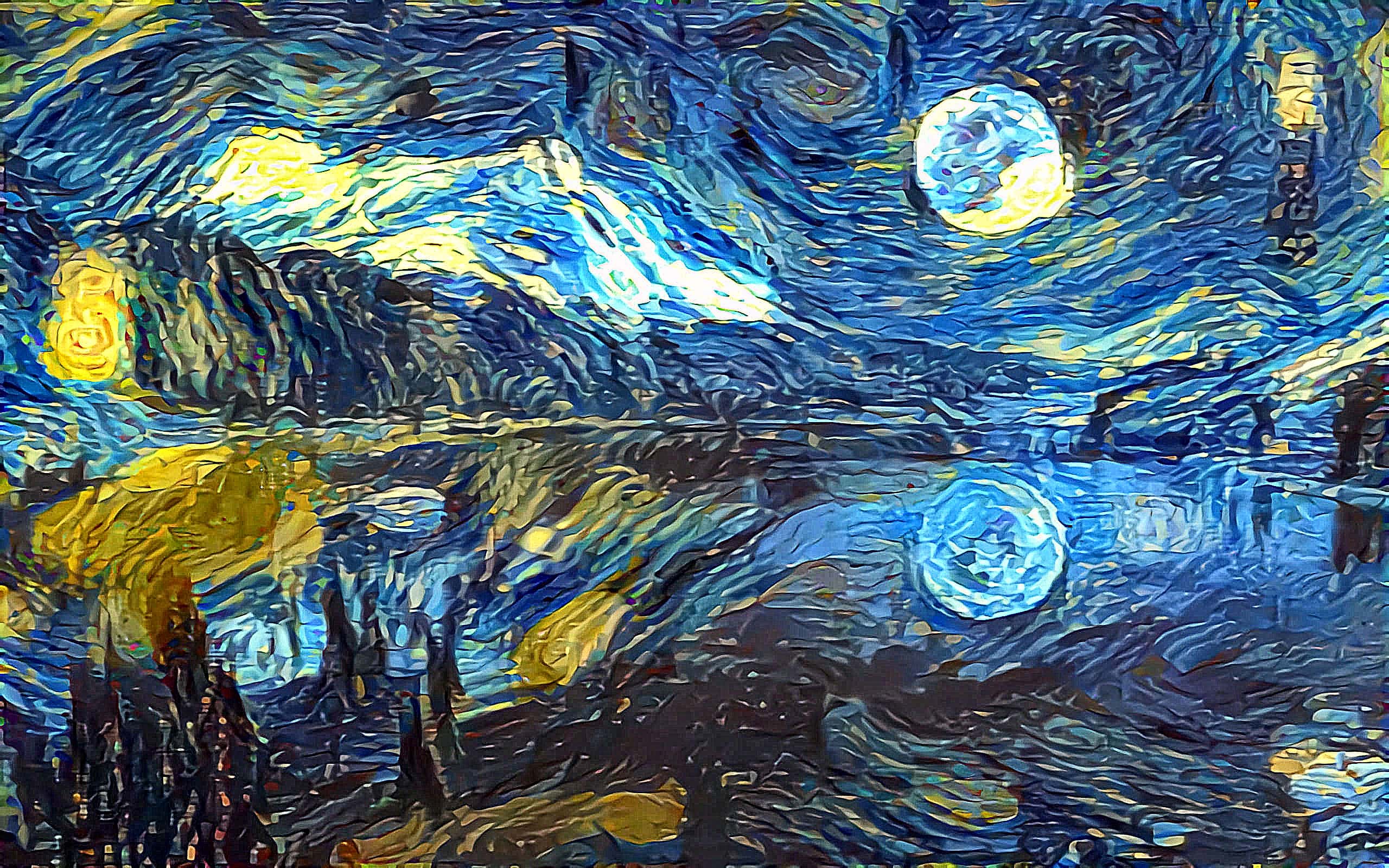} \\
(e) Cat's Eyes \& Brush Strokes + Mask & (f) Moon Overlooking Lake \& Starry Night (Van Gogh)\\[6pt]
%\multicolumn{2}{c}{\includegraphics[width=65mm]{gatys8.jpg} }\\
%\multicolumn{2}{c}{(e) fifth}
\end{tabular}
\caption{Improved Neural Style Transfer with Color + Mask Transfer}
\label{table1}
\end{figure}

The model was trained on the MS-Coco dataset \cite{lin2014microsoft} 80K training images with 1 million iterations (12.5 epochs). The total training time was 45 hours for all 5 decoders. The latter two layers took 10 and 22 hours respectively. The model was trained on a Google Cloud Platform instance with 16 Intel Skylake CPUs, 64GB RAM, and one Nvidia P100 GPU. The results of \textit{Improved Neural Style Transfer} can be observed in Figure \ref{table1}.

The above generated images in Figure \ref{table1} were up-scaled by a factor of 4 and then de-noised using \textit{Gaussian blurring} as post-processing to reduce noise from the upscaled images. We can observe that the quality of the generated images is excellent. We reiterate that this quality was obtained by using the improvements suggested by \textit{Novak et al} \cite{NovakN16improving}. %The content \ref{content_images} and style images \ref{style_images} referenced above can be seen in the appendix.

\begin{figure}[t!]
\begin{tabular}{cc}
  \includegraphics[width=65mm]{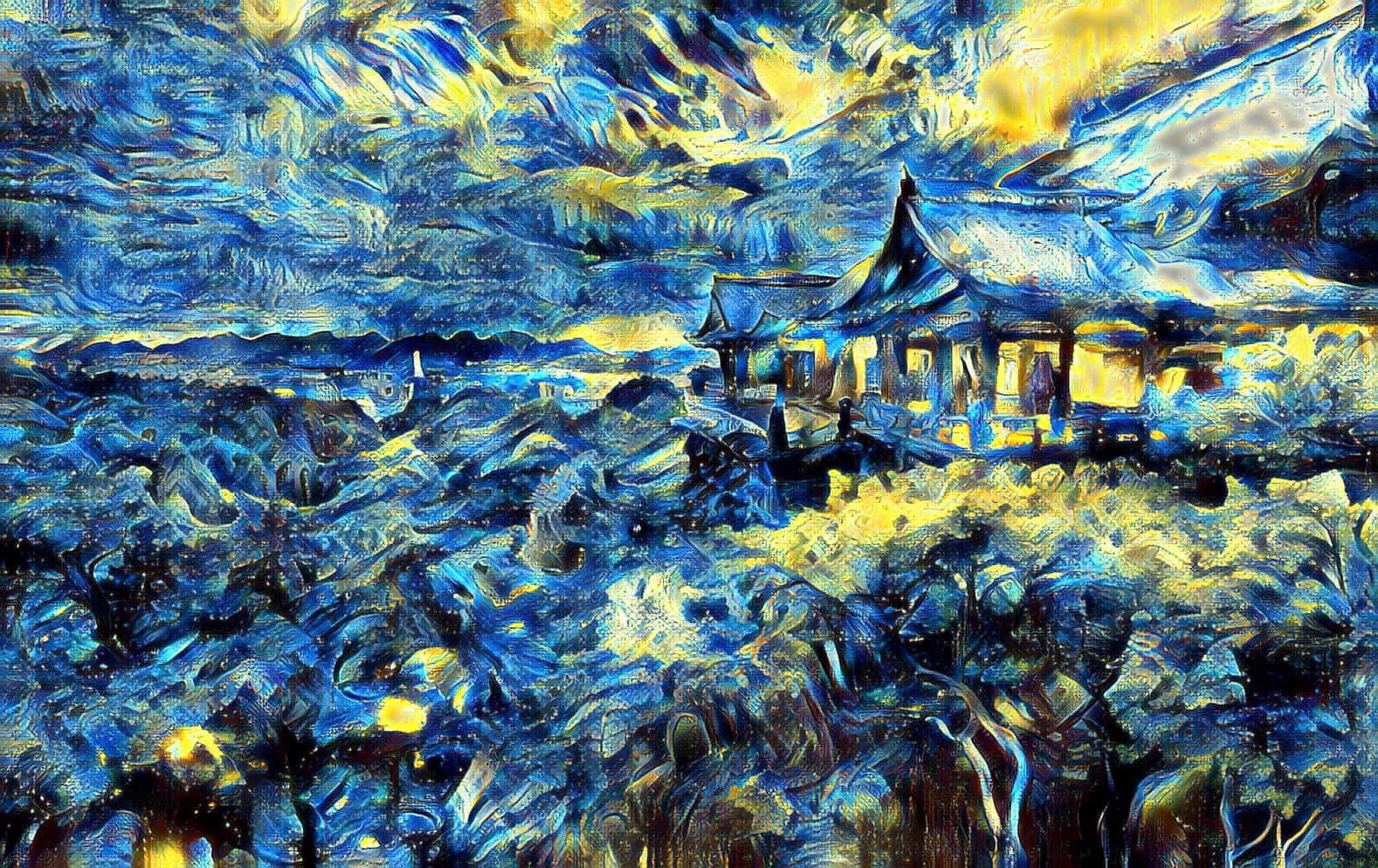} &   \includegraphics[width=65mm, height=40.5mm]{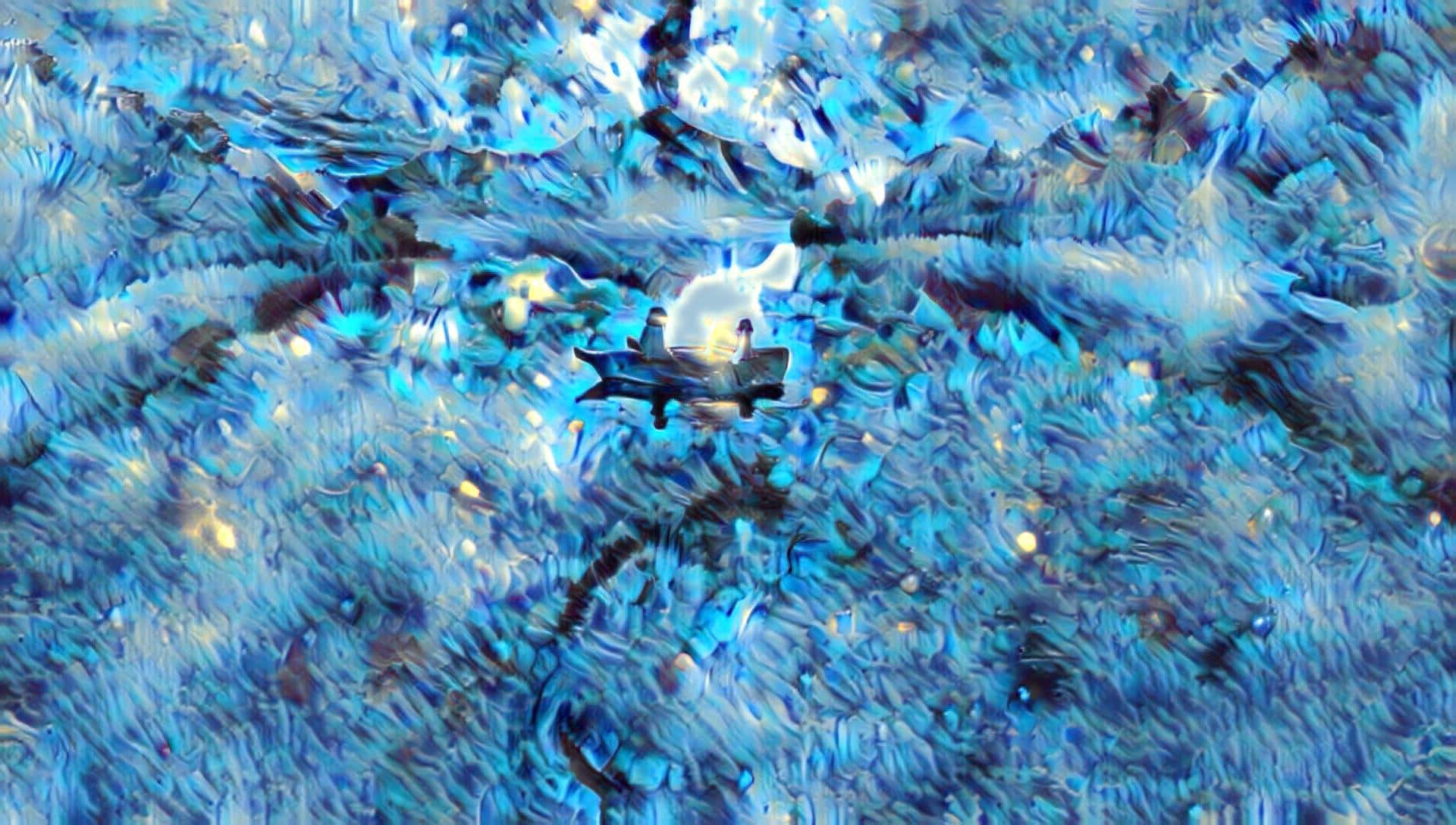} \\
(a) Japanese shrine \& Starry Night (Van Gogh) & (b) Milky Way \& Blue Strokes + Color \\[6pt]
 \includegraphics[width=65mm]{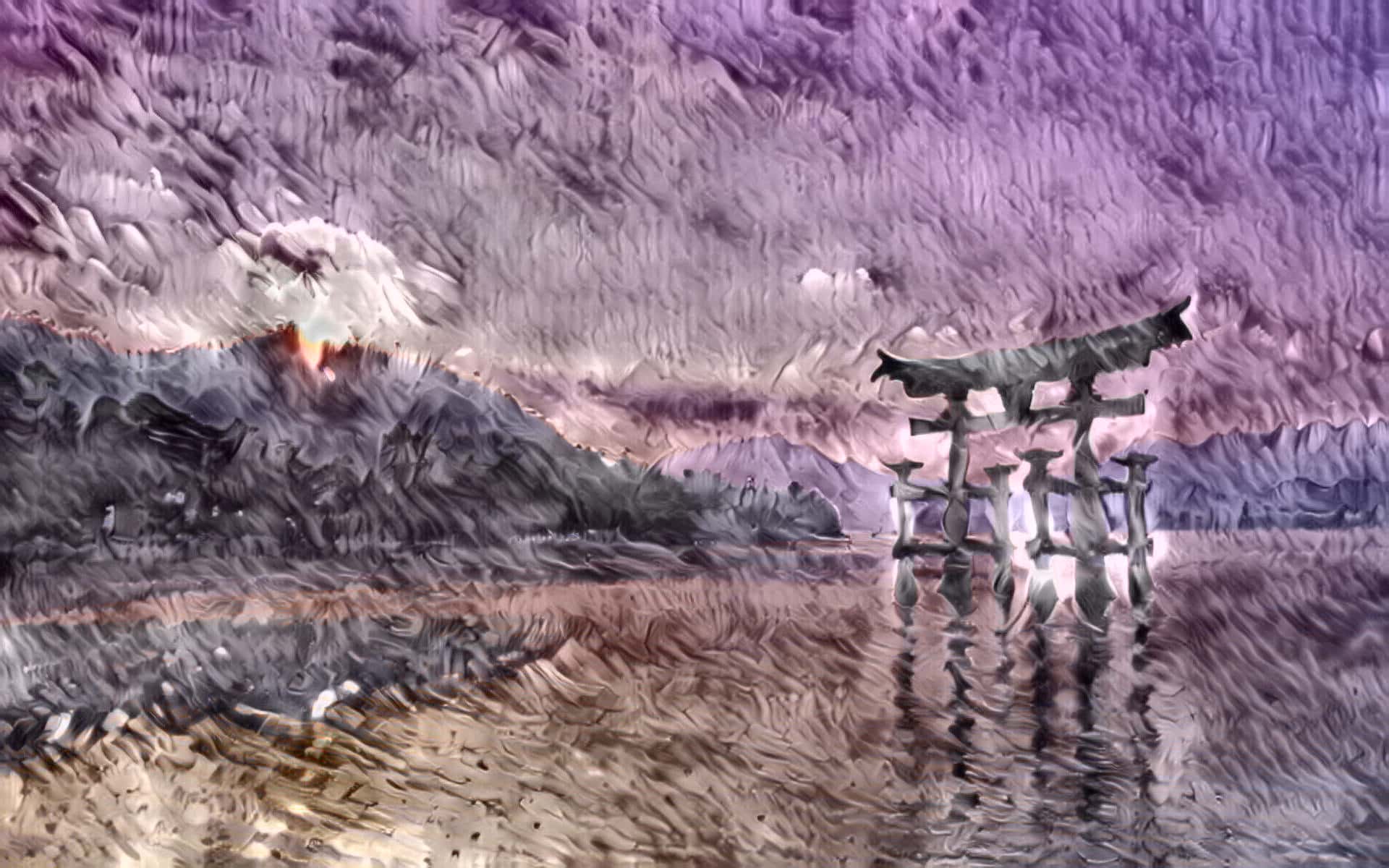} &   \includegraphics[width=65mm]{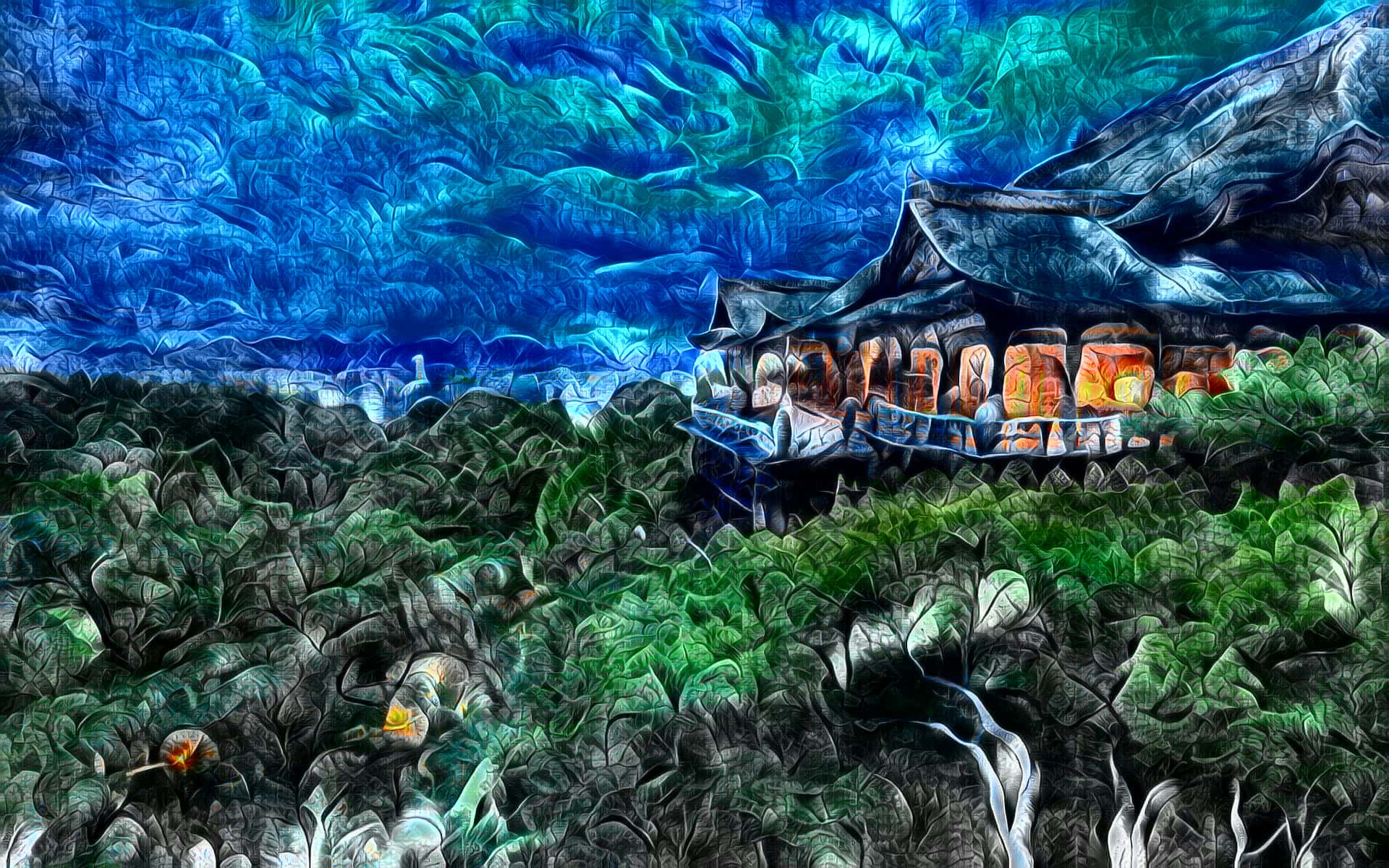} \\
(c) Itsukushima Shrine \& Blue Strokes + Color & (d) Japanese shrine \& Patterned Leaf + Color \\[6pt]
 \includegraphics[width=65mm]{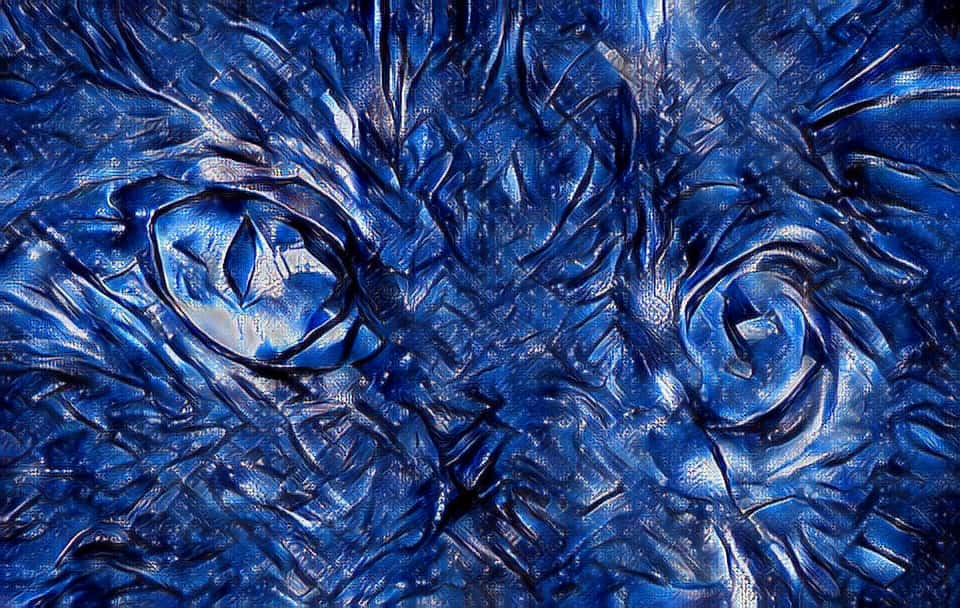} &   \includegraphics[width=65mm]{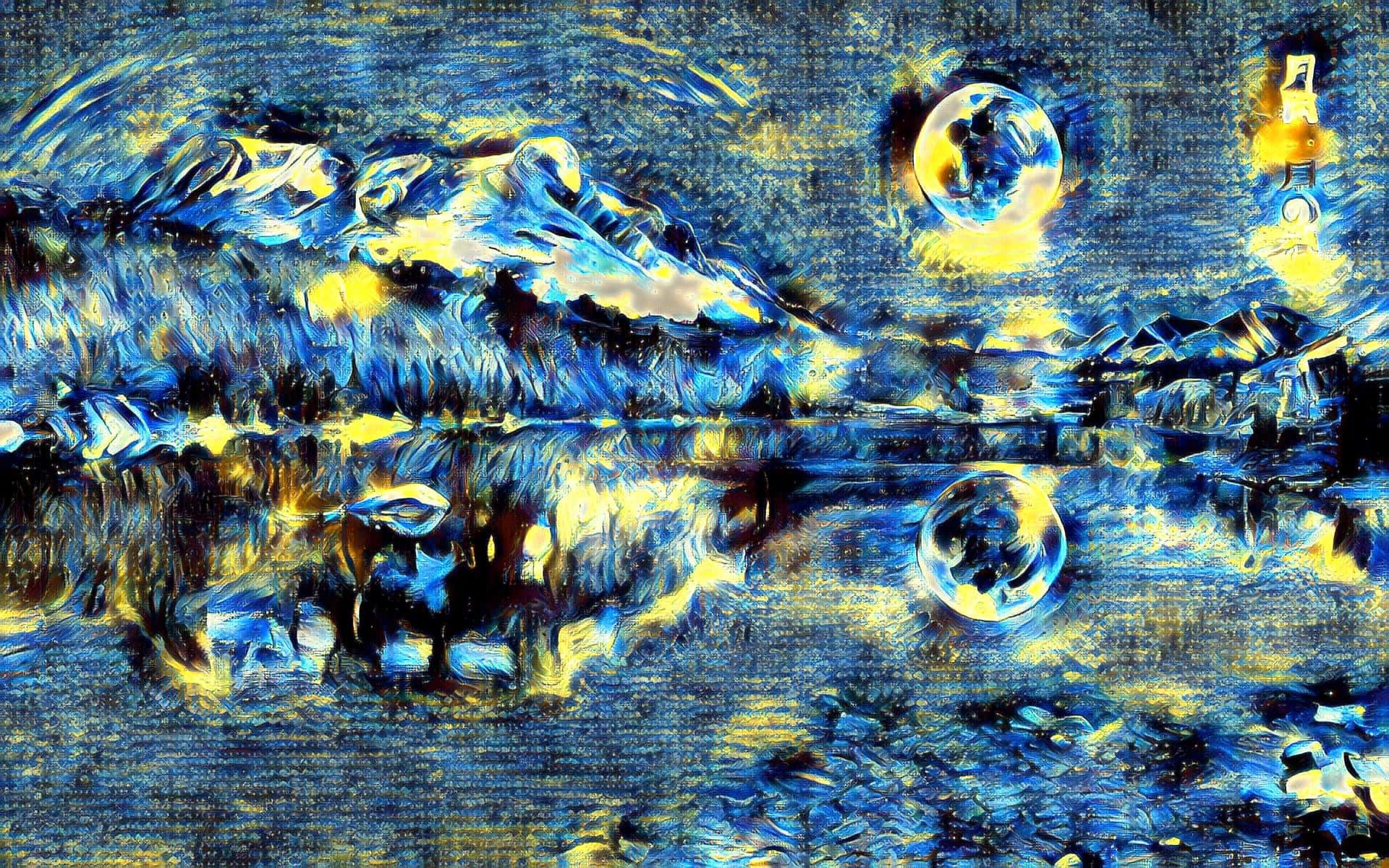} \\
(e) Cat's Eyes \& Brush Strokes + Mask & (f) Moon Overlooking Lake \& Starry Night (Van Gogh)\\[6pt]
%\multicolumn{2}{c}{\includegraphics[width=65mm]{gatys8.jpg} }\\
%\multicolumn{2}{c}{(e) fifth}
\end{tabular}
\caption{Universal Neural Style Transfer with Color + Mask Post Processing}
\label{table2}
\end{figure}

We now compare the above with the generated images obtained from the \textit{Universal Style Transfer}, which are generated at 1080p quality in less than 5 seconds each on a single GPU. Since color transfer and mask transfer cannot be obtained during the forward pass, we instead apply them as post-processing steps on the generated 1080p image. The generated images can be compared with the above in Figure \ref{table2}.

\section{Conclusion}

% Write what we learned from this project. What future work can be done to improve results.
We learned about style transfer using encoder-decoder networks. We explored the various algorithms and methods tried by previous authors and how they compare. The predominant conclusion that arises is there is a massive trade-off between speed and quality with respect to the generated images. This is clearly seen - the \textit{neural style transfer} model lets users control every aspect of tuning and training and takes a long time to train per style, but produces images with amazing quality.

The \textit{universal style transfer} model aims to alleviate some of the disadvantages of \textit{neural style transfer} by trading off some quality and introducing a general model that does not need to be fine tuned for each style image, can generate images with comparable speed, and produces visually appealing images. Users can also input larger images and get outputs that don't need rescaling or denoising using this model, unlike the \textit{neural style transfer} model. In the end, there is a huge margin for improvement in this task and much can be explored.

\subsubsection*{Acknowledgments}

We would like to thank Professor \href{https://www.cs.uic.edu/~zhangx/}{Xinhua Zhang} and \href{https://www.linkedin.com/in/vigneshganapathiraman}{Vignesh Ganapathiraman} for their invaluable knowledge, advice, and guidance.

\bibliography{biblio}

\newpage

\begin{appendices}

\newcommand{\squeezeup}{\vspace{-50mm}}

\section{Content Images}

\squeezeup

\begin{figure}[!ht]
\begin{tabular}{cc}
  \includegraphics[width=65mm, height=49mm]{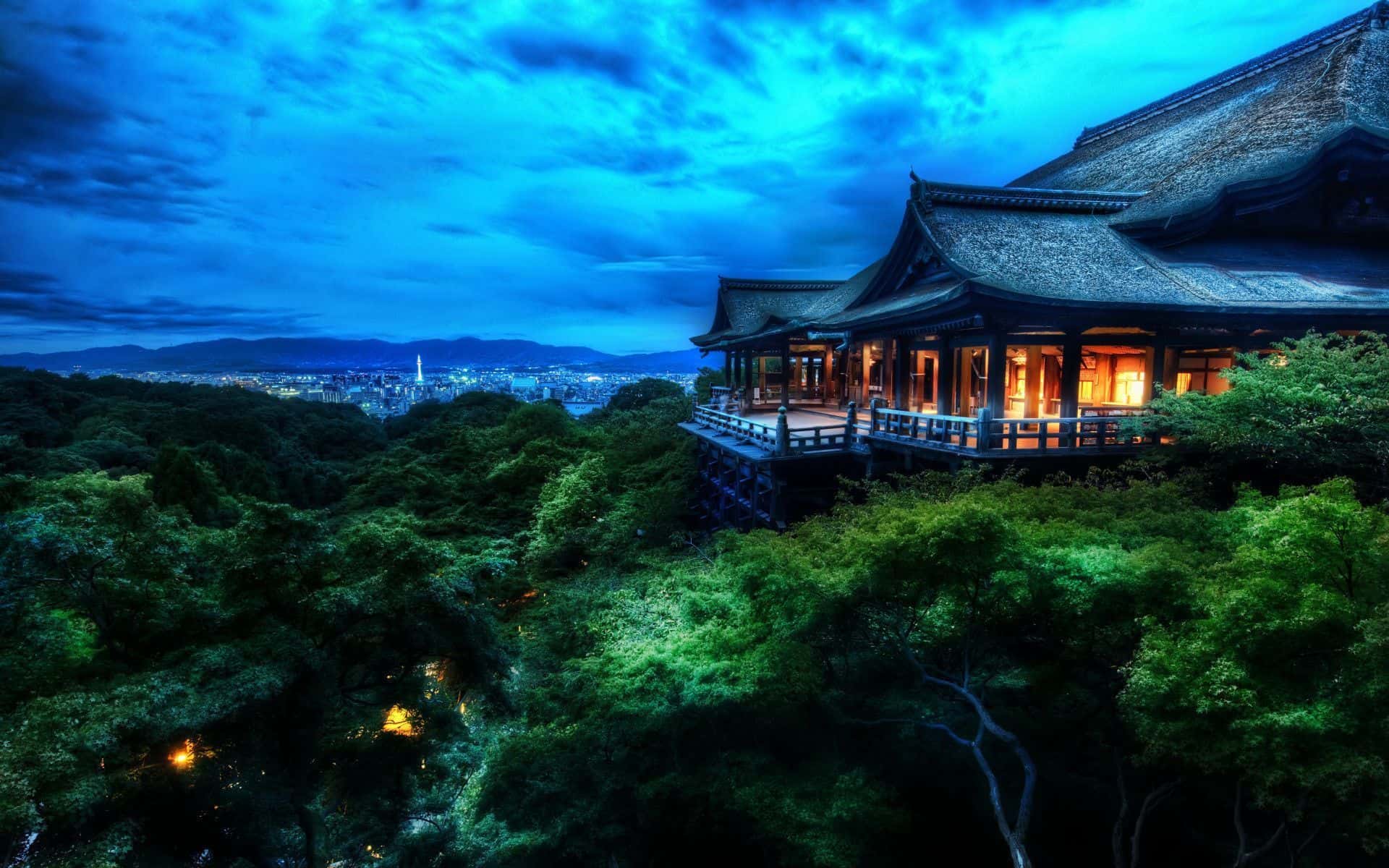} & \includegraphics[width=65mm, height=49mm]{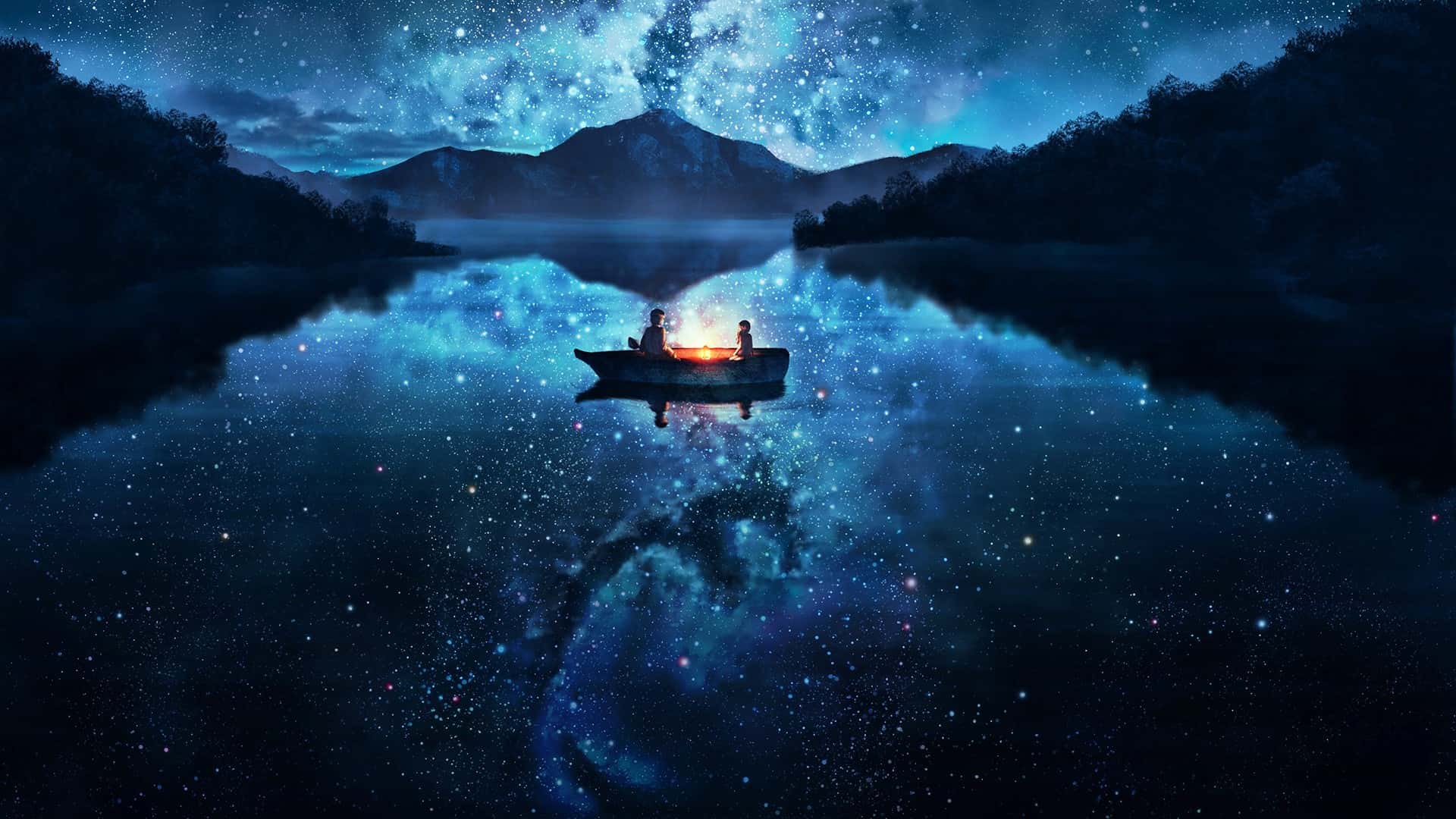} \\
(a) Japanese Shrine & (b) Milky Way \\[6pt]
 \includegraphics[width=65mm, height=49mm]{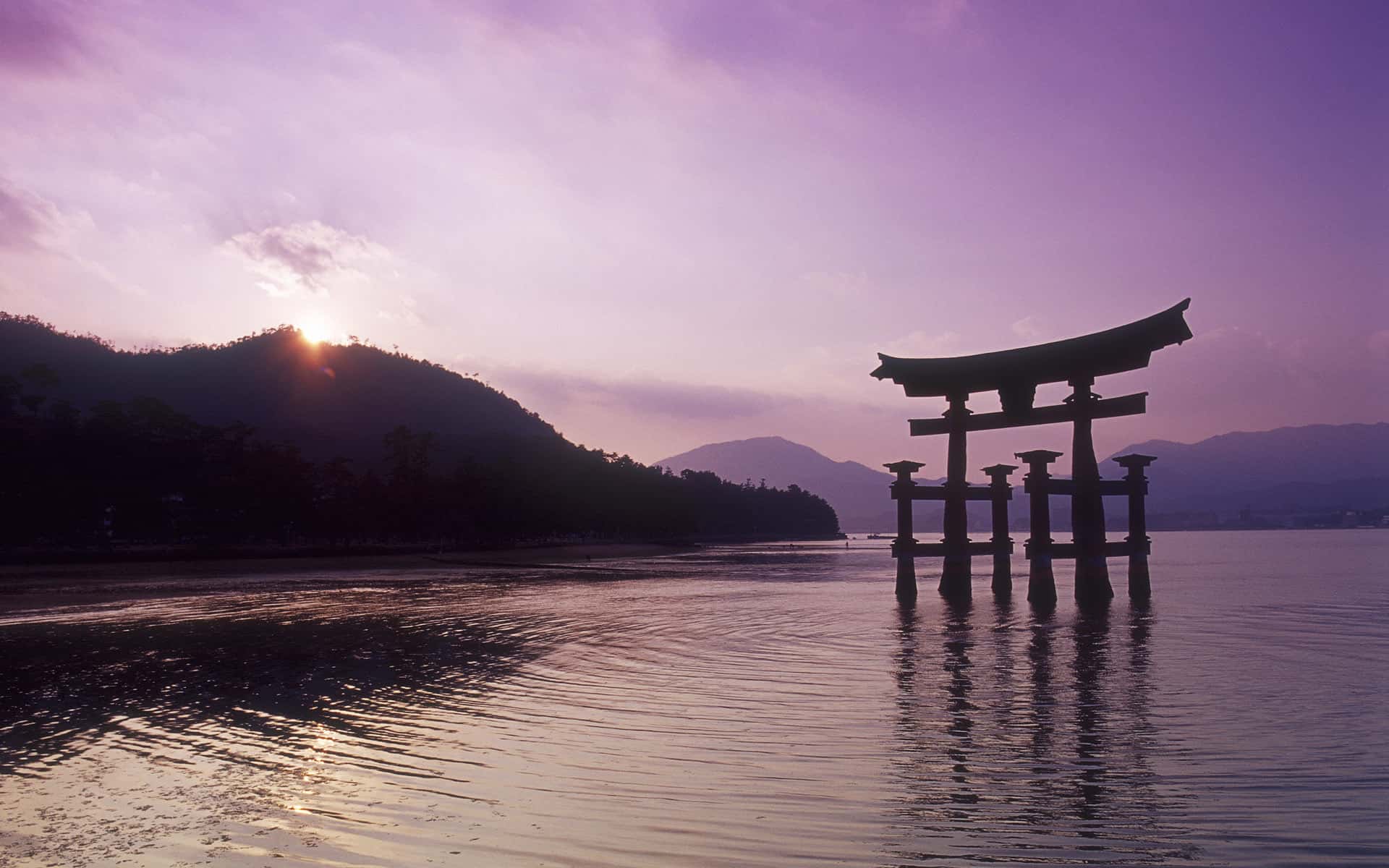} & \includegraphics[width=65mm, height=49mm]{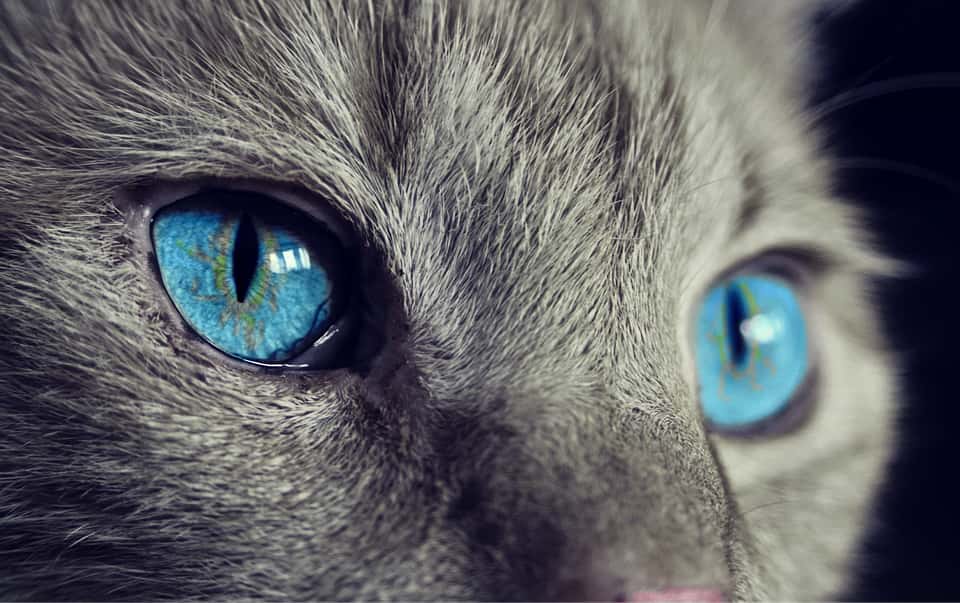} \\
(c) Itsukushima Shrine & (d) Cat's Eyes \\[6pt]
\multicolumn{2}{c}{\includegraphics[width=65mm, height=49mm]{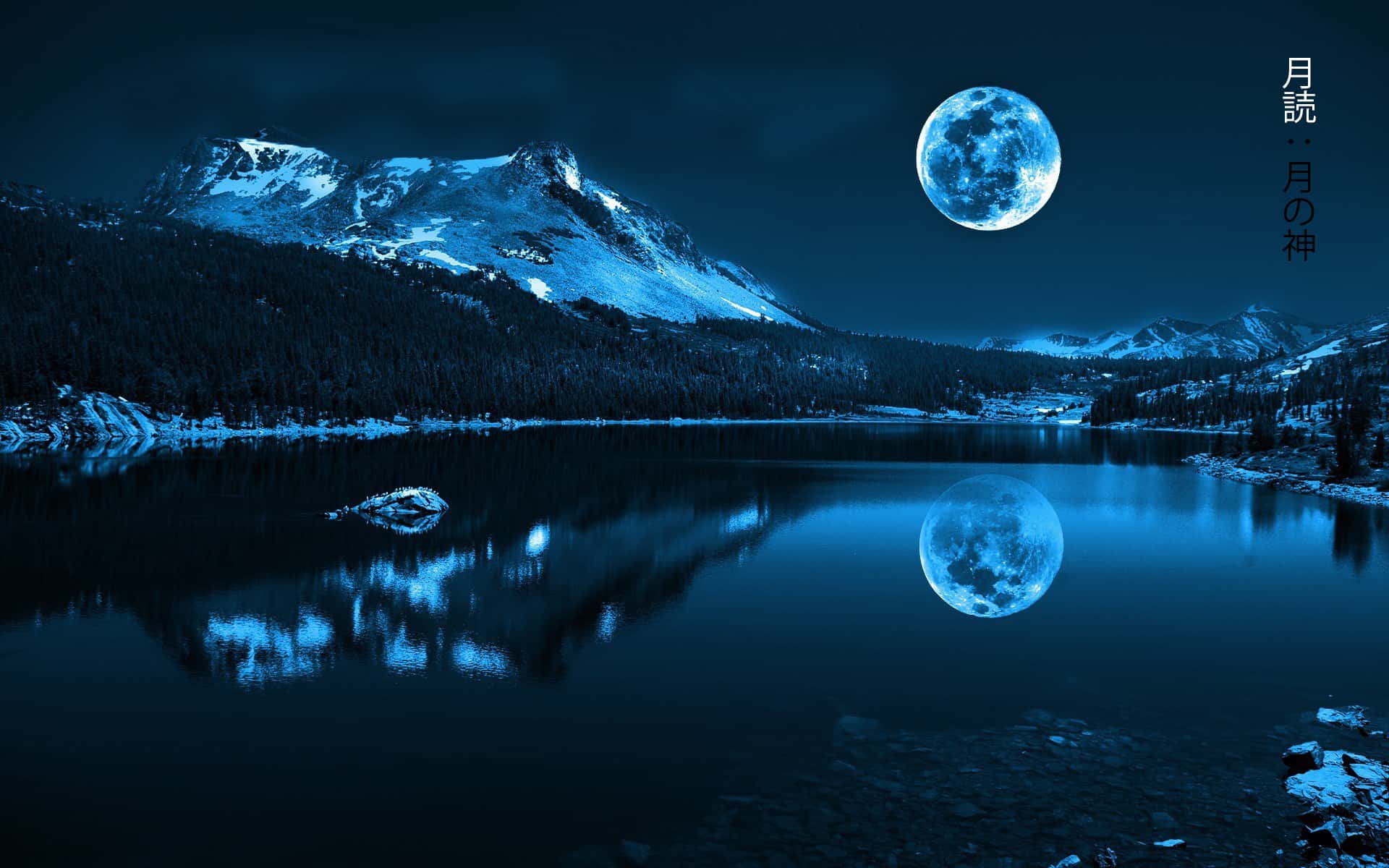} }\\
\multicolumn{2}{c}{(e) Moon Overlooking Lake}
\end{tabular}
\caption{Content Images}
\label{content_images}
\end{figure}

\section{Style Images}

\begin{figure}[H]
\begin{tabular}{cc}
  \includegraphics[width=65mm]{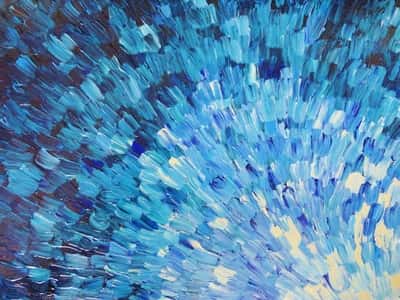} &   \includegraphics[width=65mm, height=49mm]{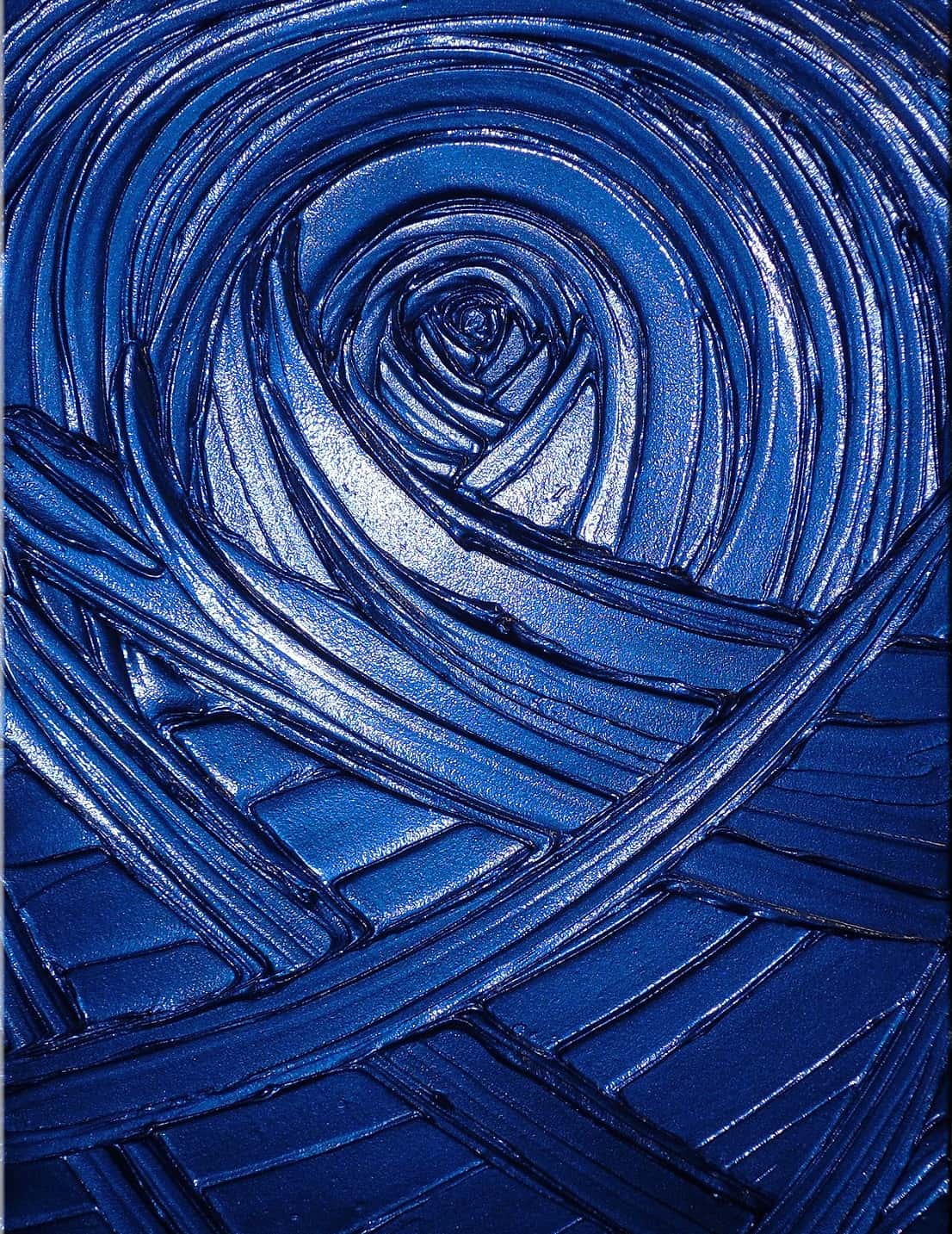} \\
(a) Blue Strokes & (b) Brush Strokes \\[6pt]
 \includegraphics[width=65mm]{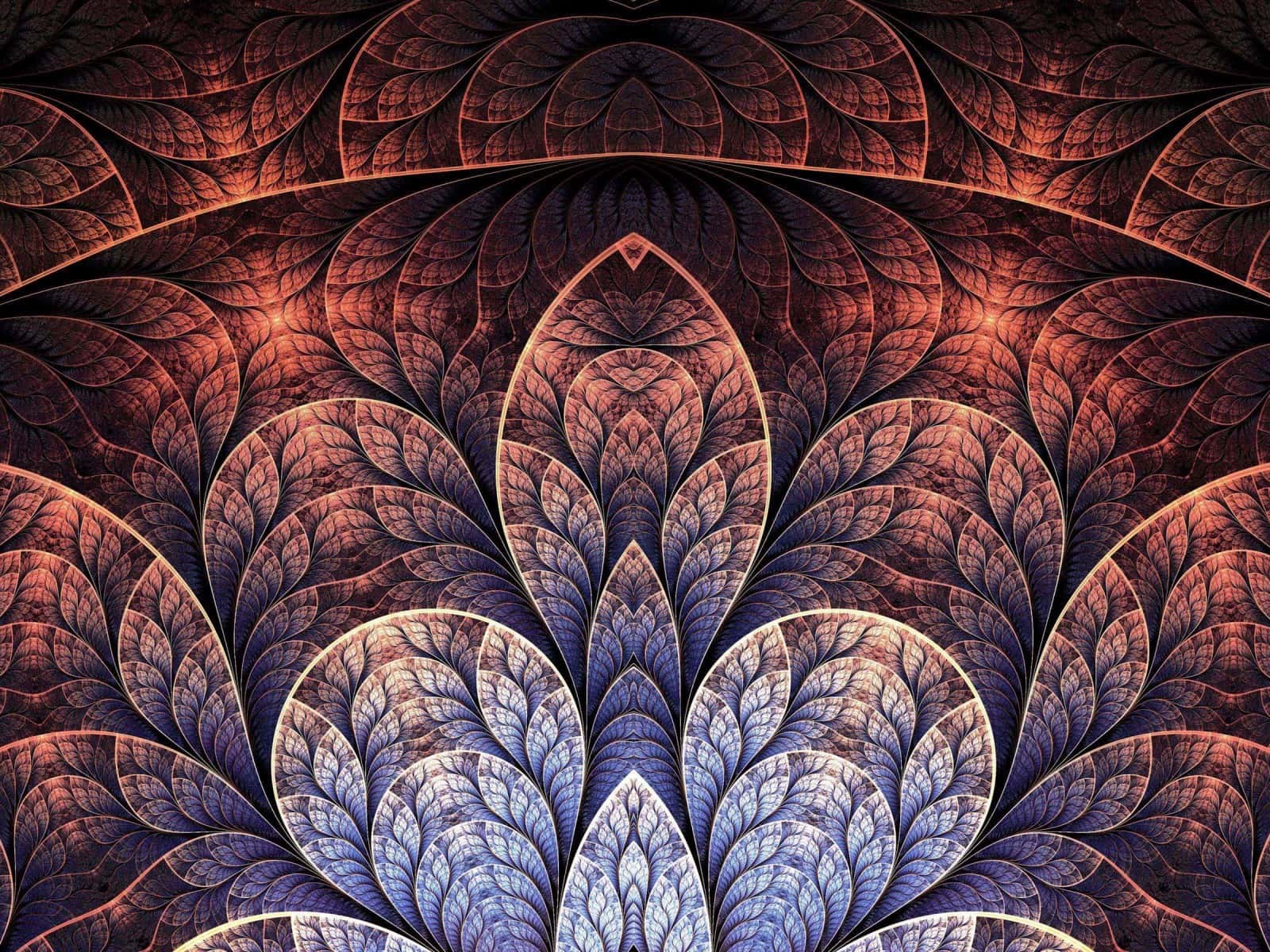} &   \includegraphics[width=65mm, height=49mm]{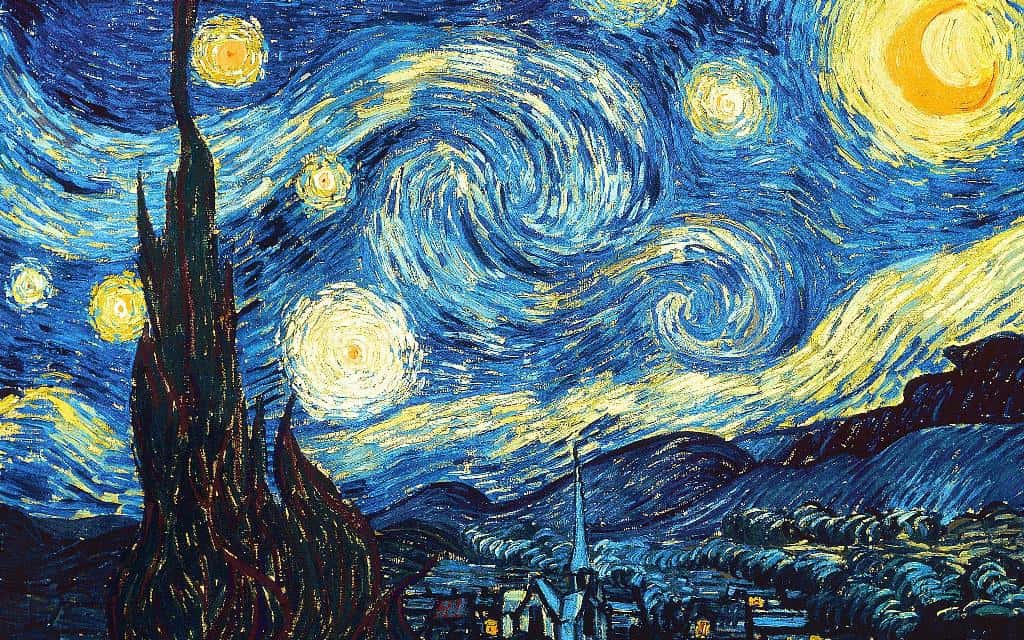} \\
(c) Patterned Leaf & (d) Starry Night \\[6pt]
%\multicolumn{2}{c}{\includegraphics[width=65mm]{gatys8.jpg} }\\
%\multicolumn{2}{c}{(e) fifth}
\end{tabular}
\caption{Style Images}
\label{style_images}
\end{figure}

\end{appendices}

\end{document}